\pdfoutput=1
\documentclass[letterpaper]{article} 
\usepackage{aaai25}  
\usepackage{times}  
\usepackage{helvet}  
\usepackage{courier}  
\usepackage[hyphens]{url}  
\usepackage{graphicx} 
\usepackage{natbib}  
\usepackage{caption} 
\usepackage{algorithm}
\usepackage{algorithmic}

\newcommand{\datasetname}{\texttt{CoMT}}

\usepackage{amsmath,amsfonts,bm}

\def\eqref#1{equation~\ref{#1}}

\def\1{\bm{1}}

\DeclareMathAlphabet{\mathsfit}{\encodingdefault}{\sfdefault}{m}{sl}
\SetMathAlphabet{\mathsfit}{bold}{\encodingdefault}{\sfdefault}{bx}{n}

\usepackage{subfigure}
\usepackage{xcolor} 
\usepackage{colortbl}
\usepackage{amsfonts} 
\usepackage{amssymb} 
\usepackage{pifont}
\usepackage{tabularx} 
\usepackage{arydshln} 
\usepackage{soul}
\usepackage{mathtools} 
\usepackage{courier}
\usepackage{adjustbox} 
\usepackage{multirow} 
\usepackage{paralist} 
\usepackage{booktabs} 
\usepackage{url}
\usepackage{enumitem}
\usepackage{tcolorbox}
\usepackage{tablefootnote}
\usepackage{gensymb}
\usepackage{bm}
\usepackage[textwidth=0.95in]{todonotes}
\definecolor{newblue}{RGB}{215,238,249}
\definecolor{my_green}{RGB}{40,154,121}
\definecolor{my_yellow}{RGB}{255,165,0}
\definecolor{my_red}{RGB}{176,46,46}
\newcommand{\correctmark}{\textcolor{my_green}{\ding{52}}} 
\newcommand{\errormark}{\textcolor{my_red}{\ding{56}}}

\usepackage{newfloat}
\usepackage{listings}
\DeclareCaptionStyle{ruled}{labelfont=normalfont,labelsep=colon,strut=off} 
\floatstyle{ruled}
\newfloat{listing}{tb}{lst}{}
\floatname{listing}{Listing}
\title{\datasetname{}: A Novel Benchmark for Chain of Multi-modal Thought on Large Vision-Language Models}
\author{
    Zihui Cheng\textsuperscript{\rm 1,\rm 2}\equalcontrib,
    Qiguang Chen\textsuperscript{\rm 3}\equalcontrib,
    Jin Zhang\textsuperscript{\rm 3},
    Hao Fei\textsuperscript{\rm 4},\\
    Xiaocheng Feng\textsuperscript{\rm 3},
    Wanxiang Che\textsuperscript{\rm 3},
    Min Li\textsuperscript{\rm 1},
    Libo Qin\textsuperscript{\rm 1,\rm 2}\thanks{Corresponding Author.},
}
\affiliations{
    \textsuperscript{\rm 1}School of Computer Science and Engineering, Central South University, China\\
    \textsuperscript{\rm 2}Key Laboratory of Data Intelligence and Advanced Computing in Provincial Universities, Soochow University, China\\
    \textsuperscript{\rm 3}Research Center for SCIR, Harbin Institute of Technology, Harbin, China\\
    \textsuperscript{\rm 4}National University of Singapore, Singapore\\
    \texttt{czh\_up@csu.edu.cn}, \texttt{qgchen@ir.hit.edu.cn}

}

\usepackage{bibentry}

\begin{document}

\maketitle

\begin{abstract}
Large Vision-Language Models (LVLMs) have recently demonstrated amazing success in multi-modal tasks, including advancements in Multi-modal Chain-of-Thought (MCoT) reasoning. Despite these successes, current benchmarks still follow a traditional paradigm with multi-modal input and text-modal output, which leads to significant drawbacks such as \textit{missing visual operations} and \textit{vague expressions}.
Motivated by this, we introduce a novel Chain of Multi-modal Thought (\datasetname{}) benchmark to address these limitations.
Different from the traditional MCoT benchmark, \datasetname{} requires both multi-modal input and multi-modal reasoning output, aiming to mimic human-like reasoning that inherently integrates visual operations.
Specifically,
\datasetname{} consists of four categories: (1) Visual Creation, (2) Visual Deletion, (3) Visual Update, and (4) Visual Selection to comprehensively explore complex visual operations and concise expression in real scenarios.
We evaluate various LVLMs and strategies on \datasetname{}, revealing some key insights into the capabilities and limitations of the current approaches. We hope that \datasetname{} can inspire more breakthroughs on introducing multi-modal generation into the reasoning process. The project page is available at \url{https://github.com/czhhzc/CoMT}.
\end{abstract}

\section{Introduction}

\begin{table*}[t]
	\centering
	\begin{adjustbox}{width=0.90\textwidth}
		\begin{tabular}{lccccccc}
			\toprule
			\textbf{Benchmark} & \textbf{\#Question} & \textbf{\#Image} & \textbf{\#VO} & \textbf{MMCoT} & \textbf{MT} & \textbf{Avg. MT Step} & \textbf{Rationale}
			\\
			\midrule
			{VCR}~\cite{zellers2019recognition} & 290k & 99,904 & \errormark & \textasciitilde4\% & \errormark & \errormark & \correctmark \\
			{A-OKVQA}~\cite{schwenk2022okvqa} & 24,903 & 23,692 & \errormark & \textasciitilde21\% & \errormark & \errormark & \correctmark \\
			{KI-VQA}~\cite{li2023comprehensive} & 4,290 & 4,189 & \errormark & \textasciitilde17\% & \errormark & \errormark & \correctmark \\
			{ScienceQA}~\cite{lu2022learn} & 21,208 & 10,332 & \errormark & \textasciitilde8\% & \errormark & \errormark & \correctmark \\
			{MMMU}~\cite{yue2023mmmu} & 11,550 & 11,264 & \errormark & \textasciitilde8\% & \errormark & \errormark & $<$18\% \\
			{M$^3$CoT}~\cite{chen2024m} & 11,459 & 11,293 & \errormark & 100\% & \errormark & \errormark & \correctmark \\
			\midrule
			{\datasetname{} (ours)} & 3853 & 14,801 & 4 & 100\%
			& \correctmark & 3.11 & \correctmark \\
			
			\bottomrule
		\end{tabular}
	\end{adjustbox}
	\caption{
		Comparison of \datasetname{} and multi-modal related datasets.\protect\footnotemark
		\textbf{\#X}: the size of X; \textbf{VO}: supported visual operations;
		\textbf{MMCoT}: the ratio of samples with multi-step MCoT (MMCoT) in the datasets;
		\textbf{MT}: Multi-modal Thought. \textbf{Avg. MT Step}: The average step of Multi-modal Thought. 
		Our benchmark has the following two advantages: (1) abundant rationale containing multi-modal thought, (2) more comprehensive and fine-grained visual operation.
	}
	\label{intro:dataset-cmp}
\end{table*}

Recently, large vision-language models (LVLMs) have achieved remarkable success across various multi-modal tasks~\citep{liu2024visual,zhu2023minigpt,qin2024large,zhang2024vision,fei2024vitron}.
In addition, LVLMs have also emerged with amazing capabilities, especially the capability of chain-of-thought (CoT) reasoning, which can perform step-by-step reasoning~\citep{lu2022learn,chen2024m,xu2024faithful,fei-etal-2023-reasoning}.
Specifically, ~\citet{zhang2023multimodal} first formally introduce the concept of Multimodal-CoT (MCoT) and extend it into a rationalizing-answering stages paradigm.
\citet{wang2024t} propose T-SciQ to distill the advanced large language models (LLMs) to smaller models for better MCoT reasoning. 
Building on this foundation, \citet{zheng2024ddcot} propose DDCoT, utilizing advanced LLMs to split questions into a series of sub-questions and then answer them by LVLMs. ~\citet{mondal2024kam} further inject the knowledge graph information into the MCoT reasoning process, reducing the hallucinations of LLMs.
\citet{he2024multi} devise a novel latent space learning approach to acquire image features through diffusion processes, achieving more complex CoT reasoning capabilities.

\begin{figure}[t]
	\centering
	\includegraphics[width=\columnwidth]{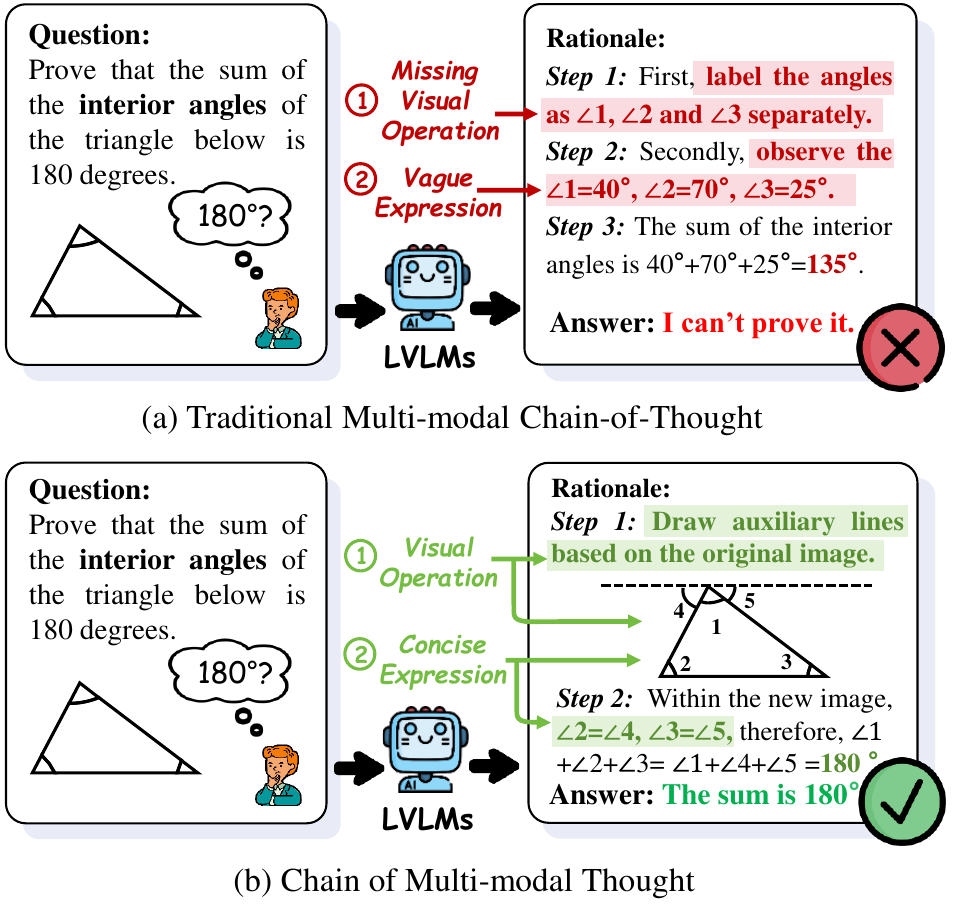} 
	\caption{Comparison between (a) traditional multi-modal CoT and (b) chain of multi-modal thought, where images in rationales are needed to be generated  from LVLMs to assist textual reasoning in rationale.} 
	\label{fig:intro-image}
\end{figure}

While remarkable success has been witnessed in MCoT, current MCoT benchmarks still follow a traditional paradigm that reads multi-modal input but can only produce single-modal reasoning output. Such a paradigm lacks integrated multi-modal reasoning output, leading to the following issues:
\setlist[itemize]{itemsep=0.1em, topsep=0.2em}
\begin{itemize}
	\item [(1)] \textbf{\textit{Missing Visual Operations}.} \textit{Effective multi-modal reasoning often requires visual operations}. However, traditional MCoT paradigms produce only textual reasoning outputs, which greatly hinders the multi-modal reasoning.
	As shown in Figure~\ref{fig:intro-image} (a), traditional methods can express operations in language, such as ``label the angles'', but they fail to execute visual operations, omitting the actual image-processing procedure.
	\item [(2)] \textbf{\textit{Vague Expressions}.} The adage ``\textit{a picture is worth a thousand words}'' highlights the limitations of text in conveying visual reasoning conditions.
	As shown in Figure~\ref{fig:intro-image} (a),  phrases like ``$\angle$1=40\degree'' are imprecise in the absence of actual annotations, failing to accurately reflect the mapping relationship between angles and measures, thus leading to ambiguity and loss of visual information.
\end{itemize}

\footnotetext{The background of traditional MCoT and our \datasetname{} can be found in our Technical Appendix A.} 

Actually, when humans perform reasoning, they naturally integrate images into the process: using visual thought for concrete, detailed reasoning while using textual thought for abstract, logical reasoning~\citep{lehmann2010core,lin2024draw,wu2024visualization}.
Take Figure~\ref{fig:intro-image} (b) as an example, LVLMs can accurately locate the specific angle by generating an annotated image. By labeling the angles and drawing auxiliary lines, LVLMs can perform clearer expressions and better multi-modal reasoning.
Inspired by this, in this paper, we aim to explore a new MCoT paradigm that requires generating  multi-modal reasoning outputs.

To fill this gap, we introduce a novel Chain of Multi-modal Thought benchmark (\datasetname{}).
Unlike the traditional MCoT benchmarks, \datasetname{} requires both multi-modal input and multi-modal reasoning output, aiming to enhance LVLMs' performance in concise expression and complex visual operations in real-world scenarios.
Specifically, \datasetname{} contains four categories to comprehensively assess the ability of LVLMs to use multi-modal thought processes:
\textbf{(1) Visual Creation} assesses the ability to generate images from scratch, thereby visualizing abstract problems; 
\textbf{(2) Visual Deletion} evaluates the removal of irrelevant information from given images;
\textbf{(3) Visual Update} examines the integration of updated images while retaining prior information;
\textbf{(4) Visual Selection} tests the selection of specific visual features for improved image comparison.
The detailed comparisons and analyses are shown in Table~\ref{intro:dataset-cmp}.

We evaluate abundant representative LVLMs and prompting strategies on \datasetname{} in extensive scenarios, yielding several \textbf{key takeaways}:
(1) \textit{\datasetname{} presents a significant challenge; nearly all zero-shot methods perform only marginally better than random, which demonstrates huge gaps compared with human performance.} (2) \textit{In-context learning (ICL) has better hope on triggering LVLMs for better multi-modal thought in \datasetname{}.} (3) \textit{Future advancements in \datasetname{} should focus on integrating multi-modal generation, logical reasoning and visual operations  into MCoT more effectively.}

Our main contributions are as follows:
\begin{itemize}
	\item To our knowledge, this is the first work to establish a benchmark for chain of multi-modal thought (\datasetname) in LVLMs, which encompasses four fundamental operations for comprehensive evaluation.
	\item We evaluate various representative LVLMs and prompting strategies, revealing a huge performance gap between LVLMs and humans. Except for Gemini, nearly all LVLMs perform at random chance levels.
	\item We explore in-context learning to enhance performance and highlight some future directions for integrating multi-modality into MCoT reasoning, hoping to provide insights for further research. 
\end{itemize}

\section{Benchmark Construction}

\begin{figure*}[t]
	\centering
	\includegraphics[width=0.91\textwidth]{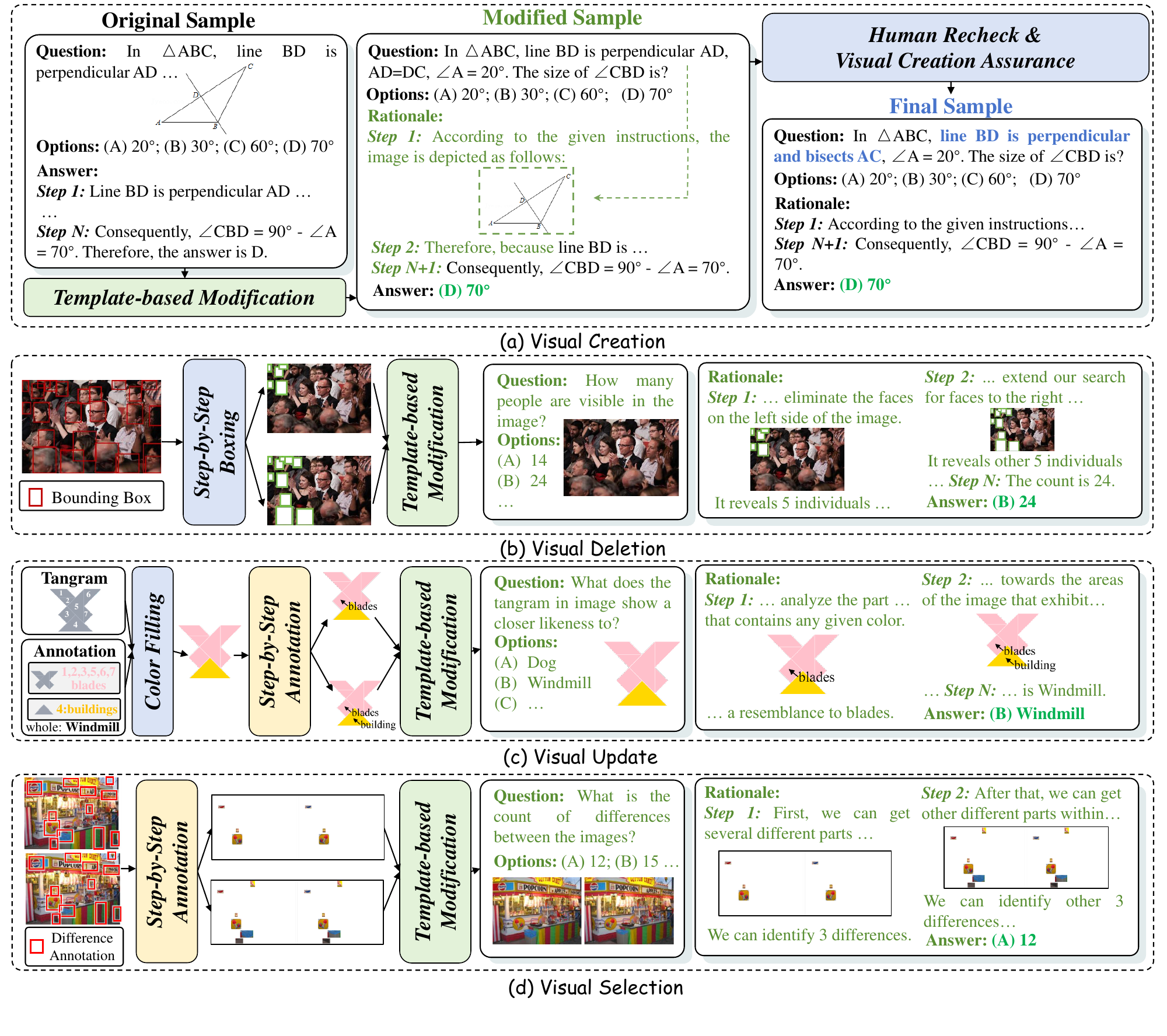}
	\caption{
		The overall annotation process for four tasks of \datasetname{}, which consists of \textit{(a)visual creation}, \textit{(b)visual deletion}, \textit{(c)visual update}, and \textit{(d)visual selection}.}
	\label{fig:overall}
\end{figure*}

\begin{figure*}[t]
	\centering
	\includegraphics[width=0.89\textwidth]{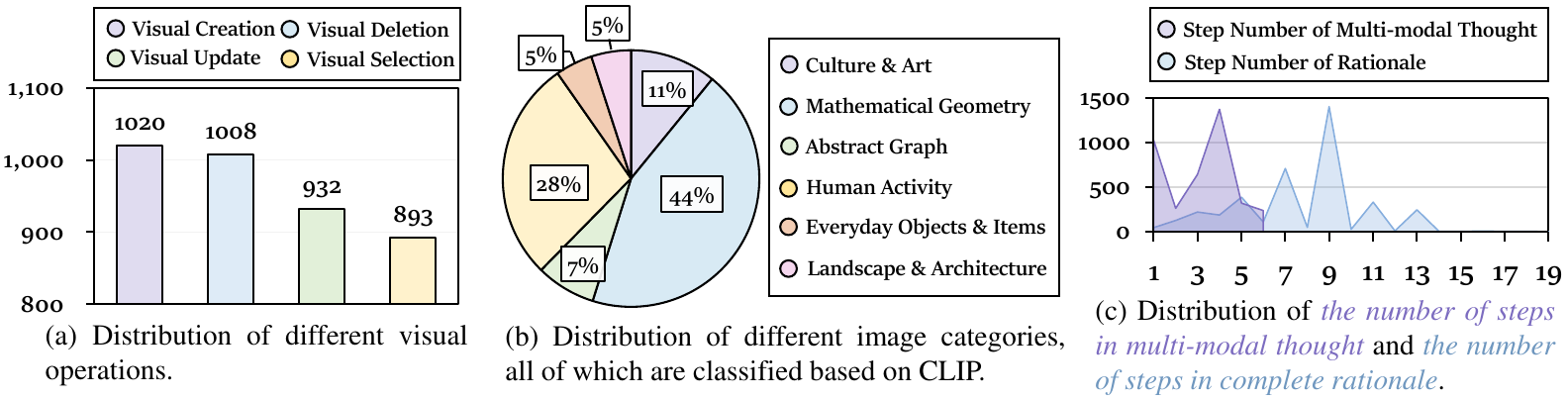}
	\caption{Distribution of \datasetname{} tasks across four types of image processing.}
	\label{fig:distribution}
\end{figure*}

We introduce \datasetname{}\footnote{The quality assurance of \datasetname{} can be found in Technical Appendix B.}, which aims to assess the ability of multi-modal thought, consisting of four types: \textit{Visual Creation} (§\ref{creation}), \textit{Visual Deletion} (§\ref{deletion}), \textit{Visual Update} (§\ref{update}), and \textit{Visual Selection} (§\ref{selection}).
Specially, we design a specified question-answering template, which involves \textit{question}, \textit{options}, \textit{image}, \textit{rationale}, and \textit{answer}, to standardize the format for all tasks within \datasetname{}.
More annotation details are shown in Technical Appendix C.

\subsection{Visual Creation}
\label{creation}
\textit{An image is worth a thousand words.}
As shown in Figure~\ref{fig:overall} (a), visual creation tasks emphasize generating images from textual descriptions to improve multi-modal reasoning.
\begin{itemize}
	\item \textit{\textbf{Original Dataset:}} We develop visual creation tasks based on the GeoQA+ dataset \citep{cao2022augmented}, which includes geometric images and textual questions as input, with textual rationales as output.
	\item \textit{\textbf{Template-based Modification:}} We first follow the template to modify the visual creation data. Specifically, we maintain the original question and option part from GeoQA+ and split the whole response into rationale and the final answer. Furthermore, we reposition the image from question to the output rationale as visual thought, with step information supplemented.
	\item \textit{\textbf{Human Recheck:}} To ensure the accurate reproduction of images, we manually augment the geometric description within the question by aligning with the image details.
\end{itemize}
 
\subsection{Visual Deletion}
\label{deletion}
\textit{In logical reasoning, it is crucial to eliminate redundant information and clarify the logical chain.} By progressively removing visual features, LVLMs experience reduced confusion, enabling step-by-step reasoning for the final answer, as illustrated in Figure~\ref{fig:overall} (b).

\begin{itemize}
	\item \textit{\textbf{Original Dataset:}} We utilize the crowd-counting task from the JHU-CROWD++ dataset~\citep{sindagi2020jhu}, which includes images with numerous faces and corresponding boxing.
	\item \textit{\textbf{Step-by-Step Boxing:}}  The most crucial aspect of crowd-counting is identifying human individuals where faces serve as a significant visual feature. To demonstrate the marking and removal of redundant visual features, we batch-mask faces based on the boxing provided, preparing for the next operation.
	\item \textit{\textbf{Template-based Modification:}} We construct the complete sample by following the \datasetname{} template, involving inquiries about the people count in the image (\textit{question}) and clarifications of the identified count (\textit{rationale}), etc. The prepared images serve as the visual thought within the rationale.
\end{itemize}

\subsection{Visual Update}
\label{update}
\textit{Marking can help sort out the logic.}
LVLMs often make mistakes in reasoning due to forgetting visual features, while humans mitigate this by annotating images. Inspired by this, as illustrated in Figure~\ref{fig:overall} (c), we propose the \textit{Visual Update} task to annotate the images step-by-step.

\begin{itemize}
	\item \textit{\textbf{Original Dataset:}} We leverage the KILOGRAM~\citep{ji2022abstract} dataset to implement tangram recognition, including the tangram image and labels of both individual pieces and the whole shape.
	\item \textit{\textbf{Tangram Annotation:}} For accurate assessments, we enhance the original tangram by applying different colors to each label category which consists of multiple individual pieces. After coloring, we explicitly annotate each category with label texts.
	\item \textit{\textbf{Template-based Modification:}} Finally, we follow the \datasetname{} template to construct the whole sample and combine the enhanced images with the textual rationales to represent the multi-modal thoughts.
\end{itemize}

\subsection{Visual Selection}
\label{selection}
\textit{Text cannot indicate the location intuitively.}
Accurately selecting among similar objects using text alone is challenging due to the inherent difficulty in precise location and difference descriptions. Following this intuition, we construct the \textit{Visual Selection} task, as shown in Figure~\ref{fig:overall} (d).

\begin{itemize}
	\item \textit{\textbf{Original Dataset:}} We construct the task from the spot-diff\footnote{https://www.allstarpuzzles.com/spotdiff/index.html} dataset. This dataset provides pairs of similar images and corresponding difference annotations, requiring precise identification of differences between two images.
	\item \textit{\textbf{Step-by-Step Annotation:}} According to the annotations, we extract the distinct areas of image pairs in batches, keeping the same position and size as the original images.
	\item \textit{\textbf{Template-based Modification:}} We then supplement the textual section within the template and integrate corresponding images to construct a multi-modal rationale.
\end{itemize}

\begin{table}[t]
	\centering
	\begin{adjustbox}{width=0.4\textwidth}
	\begin{tabular}{lc}
		\toprule
		\textbf{Statistics} & \textbf{Number}  \\
		\midrule
		Total Sample & 3,853 \\
		Total Image & 14,801 \\
		\midrule
		Average Question Length & 22.66 \\
		Average Choice Length & 1.33 \\
		Average Rationale Length & 104.74 \\
		Average Multi-modal Thought Step & 3.11 \\
		Average Rationale Step & 7.71 \\
		
		\bottomrule
	\end{tabular}%
	\end{adjustbox}
	
	\caption{Basic statistics of \datasetname{}, including sample numbers, steps of rationale, length of rationale, and image number generated in CoT.
		}
	\label{tab:statistics}
\end{table}

\begin{table*}[t]
	\centering
	\begin{adjustbox}{width=0.99\textwidth}
		\begin{tabular}{lcccccccccc}
			\toprule
			\multirow{2}{*}{Model} & \multicolumn{2}{c}{Visual Creation} & \multicolumn{2}{c}{Visual Deletion} & \multicolumn{2}{c}{Visual Update} & \multicolumn{2}{c}{Visual Selection} & \multicolumn{2}{c}{Average}
			\\\cmidrule{2-11}
			& Acc & Macro-F1 & Acc & Macro-F1 & Acc & Macro-F1 & Acc & Macro-F1 & Acc & Macro-F1
			\\
			\midrule
			Random & 27.10 & 26.75 & 25.17 & 25.15 & 24.06 & 24.05 & 25.59 & 25.55 & 25.48 & 25.37
			\\
			\midrule
			\rowcolor{gray!8}\multicolumn{11}{c}{\textit{Qwen-VL-7B}~\cite{Qwen-VL}}\\
			\midrule
			\texttt{Direct}~\cite{Qwen-VL} & 21.49 & 12.78 & \underline{26.35} & \underline{18.29} & \underline{37.64} & \underline{30.34} & 22.08 & 13.80 & \underline{26.89} & \underline{18.80} \\
			\texttt{CoT}~\cite{kojima2022large} & \underline{23.96} & \underline{19.22} & 12.63 & 11.81 & 33.62 & 26.13 & 23.22 & \underline{18.00} & 23.26 & 18.79 \\
			\texttt{Desp-CoT}~\cite{wu2023role} & 19.90 & 13.23 & 20.94 & 7.73 & 30.59 & 23.85 & \underline{26.05} & 10.48 & 24.37 & 13.82 \\ 
			\texttt{VoT}~\cite{wu2024visualization} & 22.08 & 17.51 & 14.43 & 11.71 & 28.52 & 21.02 & 22.08 & 12.47 & 21.78 & 15.68 \\ 
			\midrule
			\rowcolor{gray!8}\multicolumn{11}{c}{\textit{LLaVA-NeXT-13B}~\cite{liu2024llavanext}}\\
			\midrule
			\texttt{Direct}~\cite{liu2024llavanext} & \underline{26.34} & \underline{19.72} & 20.64 & \underline{20.06} & \underline{35.47} & \underline{34.26} & 22.76 & 19.60 & \underline{26.30} & \underline{23.41} \\
			\texttt{CoT}~\cite{kojima2022large} & 22.18 & 12.33 & 21.44 & 15.21 & 26.36 & 18.99 & 24.92 & 19.91 & 23.73 & 16.61 \\
			\texttt{Desp-CoT}~\cite{wu2023role} & 19.90 & 12.82 & 23.45 & 17.47 & 27.01 & 18.82 & 25.59 & \underline{20.77} & 23.99 & 17.47 \\ 
			\texttt{VoT}~\cite{wu2024visualization} & 20.79 & 15.58 & \underline{25.55} & 18.55 & 27.55 & 18.95 & \underline{26.61} & 17.23 & 25.13 & 17.58 \\ 
			\midrule
			\rowcolor{gray!8}\multicolumn{11}{c}{\textit{GILL}~\cite{koh2023generating}}\\
			\midrule
			\texttt{Direct}~\cite{koh2023generating} & \underline{16.93} & \underline{15.75} & \underline{22.6}5 & \underline{13.90} & \underline{23.43} & \underline{12.62} & 18.12 & \underline{10.16} & \underline{20.28} & \underline{13.11} \\
			\texttt{CoT}~\cite{kojima2022large} & 8.61 & 9.96 & 12.63 & 8.62 & 18.11 & 8.20 & 17.21 & 8.34 & 14.14 & 8.78 \\
			\texttt{Desp-CoT}~\cite{wu2023role} & 6.83 & 7.93 & 20.74 & 9.60 & 21.69 & 10.90 & \underline{20.95} & 9.12 & 17.55 & 9.39 \\ 
			\texttt{VoT}~\cite{wu2024visualization} & 5.94 & 7.01 & 17.94 & 11.81 & 21.04 & 11.51 & 14.27 & 9.23 & 14.80 & 9.89 
			\\
			\midrule
			\rowcolor{gray!8}\multicolumn{11}{c}{\textit{NExT-GPT}~\cite{wu24next}}\\
			\midrule
			\texttt{Direct}~\cite{wu24next} & \underline{24.26} & \underline{19.00} & \underline{25.75} & \underline{19.15} & 24.30 & \underline{18.04} & 22.42 & 16.24 & \underline{24.18}  & \underline{18.11} \\
			\texttt{CoT}~\cite{kojima2022large} & 20.20 & 13.88 & 23.85 & 17.25 & 22.78 & 17.95 & 21.52 & \underline{18.39} & 22.09 & 16.87 \\
			\texttt{Desp-CoT}~\cite{wu2023role} & 17.52 & 13.93 & 23.95 & 14.13 & \underline{25.38} & 17.91 & \underline{22.99} & 16.90 & 22.46 & 15.72 \\ 
			\texttt{VoT}~\cite{wu2024visualization} & 13.17 & 12.91 & 22.85 & 14.38 & 25.05 & 16.28 & 22.88 & 18.32 & 20.99 & {15.47} \\
			\midrule
			\rowcolor{gray!8}\multicolumn{11}{c}{\textit{AnyGPT}~\cite{zhan2024anygpt}}\\
			\midrule
			\texttt{Direct}~\cite{zhan2024anygpt} & 19.11 & 12.18 & 17.43 & 11.92 & 23.10 & 17.85 & \underline{27.63} & 16.91 & 21.82 & 14.72 \\
			\texttt{CoT}~\cite{kojima2022large} & 10.10 & 10.36 & 21.74 & 11.96 & 24.08 & 18.37 & 22.20 & 15.77 & 19.53 & 14.12 \\
			\texttt{Desp-CoT}~\cite{wu2023role} & \underline{19.31} & \underline{14.15} & 22.75 & \underline{12.22} & 24.84 & 18.72 & 25.59 & 16.63 & \underline{23.12} & \underline{15.43} \\  
			\texttt{VoT}~\cite{wu2024visualization} & 11.78 & 10.22 & \underline{23.45} & 11.45 & \underline{26.36} & \underline{19.44} & 25.59 & \underline{18.43} & 21.80 & 14.89 \\
			
			\midrule
			\rowcolor{gray!8}\multicolumn{11}{c}{\textit{Gemini}~\cite{team2023gemini}}\\
			\midrule
			\texttt{Direct}~\cite{team2023gemini} & 28.91 & 25.43 & \textbf{\underline{30.86}} & 22.28 & \textbf{\underline{46.36}} & \textbf{\underline{46.26}} & \textbf{\underline{27.63}} & 20.69 & \textbf{\underline{33.44}} & \textbf{\underline{28.67}} \\
			\texttt{CoT}~\cite{kojima2022large} & 27.92 & 23.07 & 28.76 & \textbf{\underline{22.73}} & 40.24 & 40.02 & 27.39 & \textbf{\underline{23.60}} & 31.08 & 27.36 \\
			\texttt{Desp-CoT}~\cite{wu2023role} & 18.04 & 14.61 & 29.36 & 21.43 & 31.05 & 23.20 & 25.14 & 11.32 & 25.90 & 17.64 \\ 
			\texttt{VoT}~\cite{wu2024visualization} & \textbf{\underline{33.27}} & \textbf{\underline{26.48}} & 27.05 & 20.79 & 35.36 & 27.83 & 24.92 & 19.38 & 30.15 & 23.62 \\
			
			\bottomrule
		\end{tabular}
	\end{adjustbox}
	\caption{
		Main results on various LVLMs. The \textbf{bold content} indicates the best performance across all models and all prompting methods, while the \underline{underlined content} signifies the best performance within a single model across all methods. See Table 4 
		in Technical Appendix F for complete results. 
	}
	\label{exp:main}
\end{table*}
\section{Benchmark Analysis}
\label{features}

\noindent \textbf{Basic statistics } As shown in Table~\ref{tab:statistics}, \datasetname{} comprises 3,853 samples and 14,801 images. \datasetname{} encompasses two primary domains within \texttt{M$^3$CoT}~\citep{chen2024m} and four visual operations (illustrated in Figure~\ref{fig:distribution} (a)) for comprehensive evaluation. Additionally, \datasetname{} requires more intricate reasoning, with an average length of 104.7 words and 7.7 steps per sample, significantly higher than ScienceQA's 48 words and 2.5 steps.

\noindent \textbf{Multi-modal diversity } 
\datasetname{} includes a diverse array of multi-modal tasks (\textit{visual creation}, \textit{visual deletion}, \textit{visual update} and \textit{visual selection}), ranging from mathematical problems to commonsense challenges, such as geometry and recognition.
Furthermore, as depicted in Figure~\ref{fig:distribution} (b), \datasetname{} features a wide range of image types encompassing ``Culture \& Art'', and ``Abstract Graph'', etc, classified by CLIP~\citep{radford2021learning}.

\noindent \textbf{Rationale diversity }
As illustrated in Figure~\ref{fig:distribution} (c), \datasetname{} exhibits a \textit{broad range} in the number of reasoning steps. Additionally, the multi-modal thought steps also show both diversity and sufficient volume. This allows for a comprehensive evaluation across different steps within \datasetname{}.

\section{Experiments}
\subsection{Experiments Setting}
In our experiments, we select a range of LVLMs as backbones, including those trained on image generation tasks as well as those that are not, including \textit{Gemini-Pro}~\citep{team2023gemini}, \textit{Qwen-VL}~\citep{Qwen-VL},
  \textit{LLaVA-NeXT}~\citep{liu2024llavanext}, 
    \textit{GILL}~\citep{koh2023generating}, \textit{NExT-GPT}~\citep{wu24next}, \textit{AnyGPT}~\citep{zhan2024anygpt}.
Additionally, we explore various prompting strategies:
(1) \texttt{Direct} prompts the model to directly generate the answer.
(2) \texttt{CoT}~\citep{kojima2022large} is a widely used prompt method to stimulate LLMs to generate steps with ``Let’s think step-by-step!''.
(3) \texttt{Desp-CoT}~\cite{wu2023role} enhances reasoning quality by instructing the model to generate a description before answering.
(4) 
\texttt{VoT}~\cite{wu2024visualization} utilizes ``Visualize the state after each reasoning step.'' to imagine the reasoning path with text-modal. 
Following \citet{qin2023cross} and \citet{chen2024m}, we extract the final generated answers using regular expressions.
See Technical Appendix D
for further experimental details.

\subsection{Main Results}
Table \ref{exp:main} presents the main results, from which we derive the following key findings:

\noindent \textbf{\textit{All LVLMs perform poorly on the \datasetname{}.}}
Despite Gemini achieving a 28.67\% F1 score across four tasks, this performance is marginally better than the random baseline by 3.3\%, indicating significant room for improvement. Additionally, except for Gemini,
most models perform at or below random levels. We attribute these to the lack of multi-modal reasoning in current LVLMs.

\noindent \textbf{\textit{Traditional Multimodal CoT almost completely fails on \datasetname{}.}} We observe that pure text-modal CoT does not attain improvement in addressing the \datasetname{} problem and even degrades the performance of most models to near-random levels. We attribute it to the fact that the inability of the model to execute specific visual logic expressions and operations results in poor performance.

\noindent \textbf{\textit{All models fail to visualize thought in textual words.}} As demonstrated in Table \ref{exp:main}, all LVLMs fail to utilize \texttt{VoT} effectively to improve performance.
Specifically, \texttt{VoT} prompts LVLMs to visualize states through \textit{textual} representation and results in an average accuracy decrease of 12.28\%. This finding suggests that although textual representation can convey visual features, the inherent differences between modalities still constrain the expression of multi-modal thought.

\subsection{Analysis}
This section will conduct a further analysis on \datasetname{}.
See Technical Appendix E 
for more implementation details.

\begin{figure}[t]
	\centering
	\includegraphics[width=0.87\linewidth]{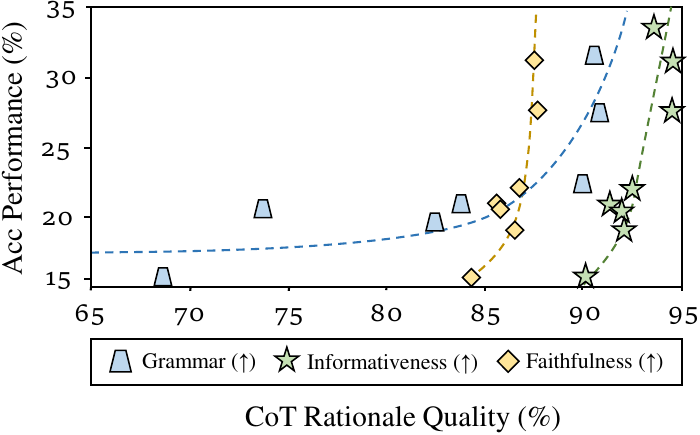}
	\caption{
		Analysis of the correlation between the model performance and the quality of rationale for different LVLMs based on ROSCOE~\citep{golovneva2023roscoe}. }
	\label{exp:rationale-quality}
\end{figure}

\noindent \textbf{\textit{Improving the quality of rationale is essential for  \datasetname{}.}}
As illustrated in Figure~\ref{exp:rationale-quality}, the quality of CoT rationale significantly impacts the \datasetname{} performance. Poor rationale quality constrains the logical coherence of LVLMs, limiting their reasoning capacities, which aligns with \citet{chen2024m}. Consequently, enhancing reasoning quality in LVLMs is a crucial area for further exploration.

\begin{figure}[t]
	\centering
	\includegraphics[width=0.95\linewidth]{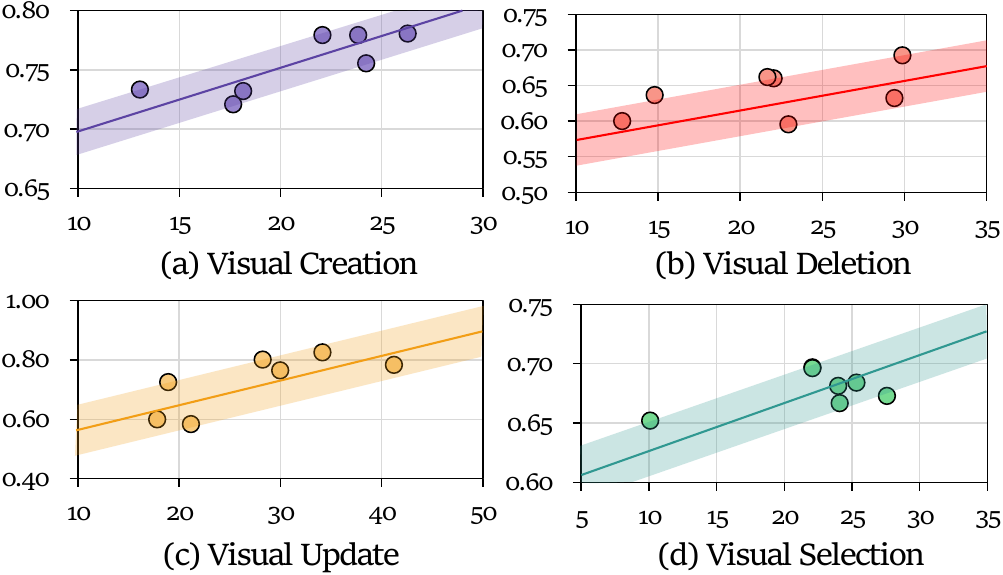}
	\caption{
		CLIPScore of LVLMs on 4 tasks within \datasetname{}. The x-axis represents the \textit{CLIPScore}, and the y-axis represents the \textit{accuracy}. }
	\label{exp:clipscore}
\end{figure}

\noindent\textbf{\textit{\datasetname{} benefits from improved multi-modal thought.}} To assess the impact of multi-modal thought on performance within \datasetname{}, we calculate the CLIPScore~\citep{hessel2021clipscore} to reflect the similarity between model output and each image within the ideal rationale pre-defined. Averaging these scores yields a multi-modal alignment score for each reasoning chain generated. As shown in Figure~\ref{exp:clipscore}, there is a significant positive correlation between performance and multi-modal alignment scores across four tasks, which indicates that \datasetname{} benefits from more multi-modal thought.

\begin{figure}[t]
	\centering
	\includegraphics[width=\linewidth]{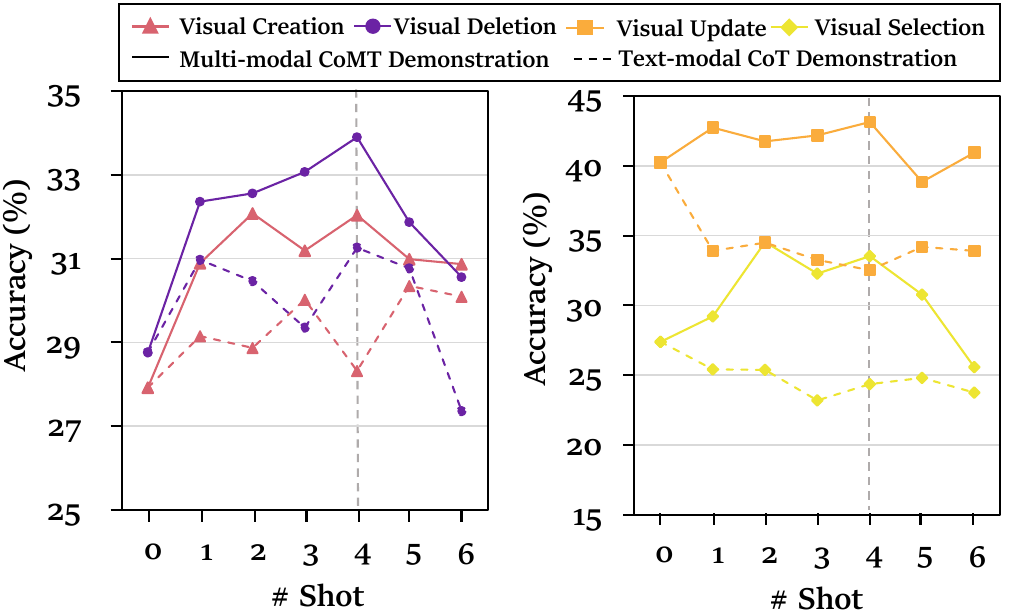} 
	\caption{Analysis on In-context Learning of \textit{Gemini-Pro}~\citep{team2023gemini} in \datasetname{}.} 
	\label{exp:few-shot}
\end{figure}

\noindent\textbf{\textit{The performance relies more on the quality of multi-modal alignment than on parameter size.}}
As shown in Table 4 
in Technical Appendix F, the IDEFICS2-8B, with fine-grained multi-modal alignment, surpasses the 13B
models, even approaching the performance of the Gemini-Pro ($>$100B,~\citet{team2023gemini}). We think that \datasetname{} performance depends more on multi-modal alignment quality rather than parameter size.

\subsection{In-context Learning Explorations}

\noindent\textbf{\textit{In-context Learning with multi-modal input and output can effectively promote the performance in \datasetname{}.}}
As shown in Figure~\ref{exp:few-shot}, using in-context learning (ICL)~\cite{li2023unified,qin2024factors} with multi-modal input and multi-modal output demonstrations significantly improves performance. It not only surpasses zero-shot prompting but also outperforms ICL with text-modal output. This approach can be successful due to the fact that LVLMs can learn to effectively facilitate  multi-modal thought through such demonstrations, even though Gemini is limited to producing rationales in the textual modality alone.

\noindent\textbf{\textit{Not more demonstrations means better performance in \datasetname{}.}}
As shown in Figure~\ref{exp:few-shot}, the model exhibits a significant downward trend in performance when the number of demonstrations exceeds \textit{four}. It shows that more demonstrations are not necessarily better, as multimodal demonstrations often require the consumption of a substantial number of tokens, which can also lead to more complex challenges associated with longer contexts.

\subsection{Error Analysis}
\begin{figure}[t]
	\centering
	\includegraphics[width=0.85\linewidth]{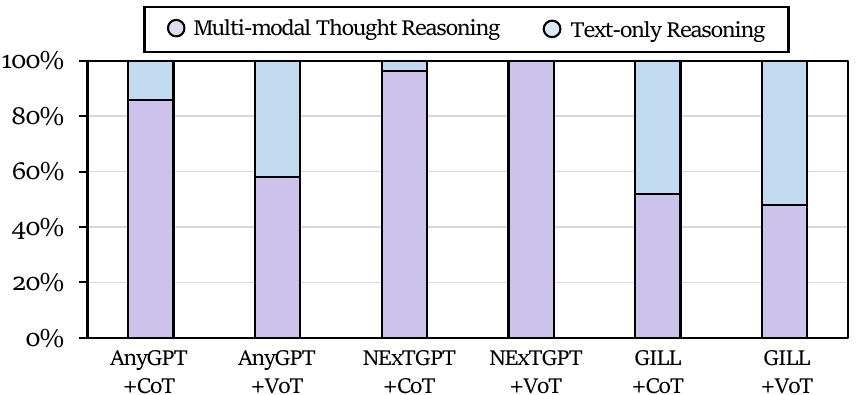} 
	\caption{Image generation frequency during reasoning.} 
	\label{exp:image-gen}
\end{figure}

\begin{figure}[t]
	\centering
	\includegraphics[width=0.85\linewidth]{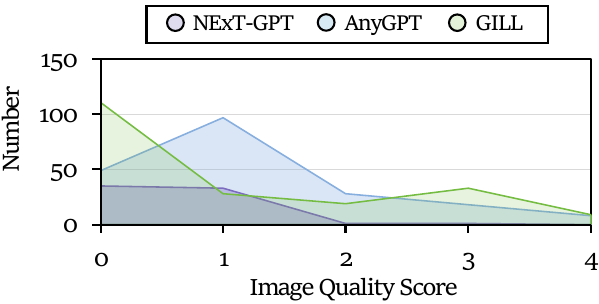} 
	\caption{Distribution of human-evaluated image quality scores ($\uparrow$) which are mainly determined based on \textit{Relevance} and \textit{Logical Correctness}. See Technical Appendix C.3 
		for evaluation details.
	} 
	\label{exp:image-distribution}
\end{figure}

\paragraph{Insufficient Multi-modal Thought.}
When dealing with multi-modal problems, models struggle to integrate multi-modal thought most of the time. As illustrated in Figure~\ref{exp:image-gen}, we observe that despite certain models (e.g., GILL, NExT-GPT, AnyGPT) being trained on image generation tasks, at least 48\% of their reasoning processes do not incorporate image generation. This occurs even when image generation is crucial for accurate outcomes, indicating a disjunction between image generation and text processing.
\paragraph{Inaccurate Textual Reasoning.}
When logical errors occur in textual reasoning, they hinder the advancement towards the correct answer. For example, Figure 10 in Technical Appendix 
 reveals that the model demonstrates poor reasoning logic, with significant logical errors, such as calculation mistakes (like  ``2*5*5=2*10''). These inaccurate textual reasoning significantly impedes progress in this field.
\paragraph{Incoherent Visual Reasoning.}
Although certain models generate images when reasoning, not all image contents align with the reasoning path, revealing an immature interaction between modalities.
We manually evaluate the generated images, with results shown in Figure~\ref{exp:image-distribution}. The distribution reveals that current LVLMs often generate \textit{irrelevant} images during reasoning (an average of 43\%, represented by \textit{score 0}) and fail to perform effective visual logic (on average 45\% of images exhibit \textit{logical mistake}, represented by \textit{score 1,2}). The judgment criteria can be found in Technical Appendix C.3. 
To be specific, Figure 11 
in Technical Appendix G 
shows instances with irrelevant text and image logic. 

\subsection{Future Directions}
Based on the above analysis, we summarize the future directions for current LVLMs tackling \datasetname{}.

\noindent\textbf{\textit{How can we effectively integrate multi-modal thought reasoning?}} 
The absence of visual thought significantly increases the difficulty when addressing certain multi-modal tasks, such as \datasetname{}. How to enable models to integrate multi-modal reasoning is an intriguing research topic. Furthermore, given the inherent differences between textual and visual modalities, exploring how to align these two modalities during reasoning presents another valuable challenge.

\noindent\textbf{\textit{How can we enhance logical reasoning capabilities for textual reasoning?}}
The inadequacies in textual reasoning logic lead to inaccurate conclusions during inference, such as calculation mistakes. Therefore, how to enable models with better textual logic to perform effective text reasoning is a critical topic to explore.

\noindent\textbf{\textit{How can we achieve effective vision logic for visual reasoning?}}
Since some generated images fail to perform effective visual logic or even be irrelevant, not all visual thoughts generated have a positive influence on the reasoning. How to enable models to develop better visual logic to produce images that are relevant and consistent with the progression of rationale is a topic worth exploring.


\section{Related Work}
The emergence of Multi-modal Chain-of-Thought (MCoT) techniques elicits the step-by-step zero-shot and few-shot multi-modal reasoning capabilities of Large Vision-Language Models (LVLMs)~\citep{wangtowards,wang2024enhancing,chen-etal-2024-rg,chen2024internvl,liu2023retrieval,he2024multi,qin2024factors,fei2024video,fei2024enhancing}.
Pioneering work introduces the ScienceQA benchmark~\citep{lu2022learn}, involving multimodal scientific questions. Subsequently, \citet{zhang2023multimodal} formally propose the concept of MCoT and introduce a two-stage framework encompassing both reasoning and answering.
Additionally, \citet{tan2024retrieval,wang2024t,zhang2024cocot,mondal2024kam,lee2024multimodal} introduce more knowledge to improve the performance and reduce hallucinations in MCoT reasoning.
Following this, \citet{zheng2024ddcot} propose DDCoT, which breaks down the question into a series of sub-questions and solves them using LVLMs.
Building upon this, \citet{chen2024m} further introduce a multi-domain multi-step multi-modal benchmark to fully evaluate the complex MCoT capabilities.
Based on traditional MCoT, some works begin preliminary exploration integrating the diffusion model or retriever model as a tool for better MCoT. \citet{meng2023chain} propose CoI to generate images as intermediate reasoning steps in single modal tasks, outperforming purely textual CoT. 
\citet{wu2024visualization} propose VoT, requiring text-only LLMs to imagine their vision reasoning paths, which increases the spatial reasoning abilities.

In contrast to our work, their strategies rely solely on textual modalities for reasoning, lacking visual operation or detailed visual expression in reasoning. To fill this gap, we propose \datasetname{} to comprehensively reveal diverse multi-modal thought capabilities. We hope \datasetname{} will inspire research on 
promoting better multi-modal reasoning.

\section{Conclusion}
In this work, we introduce a Chain of Multi-modal Thought (\datasetname{}) benchmark to evaluate and improve the multi-modal reasoning capabilities of Large Vision-Language Models (LVLMs). Through extensive experiments, our findings reveal a significant performance gap between LVLMs and human, with models generally not outperforming random chance in zero-shot scenarios. In-context Learning with multi-modal rationale emerges as a promising approach to better integrate visual and textual reasoning in LVLMs. We hope this research lays the groundwork for future enhancements in multi-modal reasoning technologies.

\appendix

\section*{Acknowledgments}
This work was supported by the National Natural Science Foundation of China (NSFC) via grant 62306342, 62236004, 62441603 and 62476073. This work was also sponsored by the Excellent Young Scientists Fund in Hunan Province (2024JJ4070), the Science and Technology Innovation Program of Hunan Province under Grant 2024RC3024 and CCF-Zhipu Large Model Innovation Fund (NO.CCF-Zhipu202406)). 
We are grateful for resources from the High Performance Computing Center of Central South University. Libo Qin is the corresponding author.

\bibliography{aaai25}

\newpage

\renewcommand{\thefigure}{9}
\begin{figure}[t]
	\centering
	\includegraphics[width=0.99\linewidth]{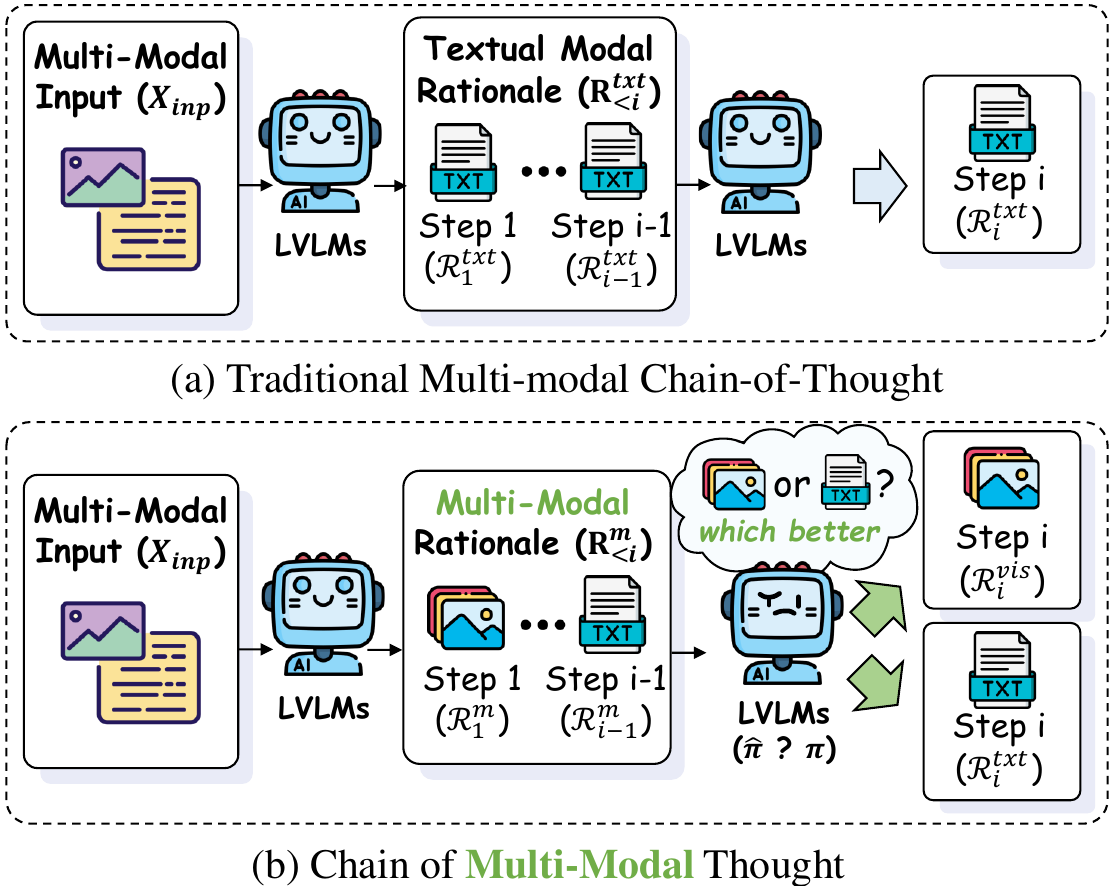}
	\caption{
		Comparison of paradigm between (a) traditional multi-modal chain of thought and (b) chain of multi-modal thought.}
	\label{fig:background_comparison}
\end{figure}

\section{Background}

\subsection{Multi-modal Chain-of-Thought}
\label{text}
As shown in Figure~\ref{fig:background_comparison} (a), the traditional multi-modal Chain-of-Thought (MCoT) involves generating a text-modal rationale, based on a multi-modal input  $\mathcal{X}_{inp}=[\mathcal{X}_{txt}; \mathcal{X}_{vis}]$. The LVLM generates a textual rationale step $\mathcal{R}^{txt}_i$ based on the rationales from the previous $i-1$ steps rationales $\mathbf{R}^{txt}_{<i}$. This process can be mathematically represented as:
\begin{equation}
	\mathcal{R}^{txt}_i=\underset{\mathcal{R}^{txt}}{\operatorname{argmax}}\  \pi(\mathcal{R}^{txt}|\mathcal{X}_{inp}, \mathbf{R}^{txt}_{<i}),
\end{equation}
where $\pi(\cdot)$ denotes the probability of the model generating the rationale $\mathcal{R}^{txt}$ from the vocabulary of textual tokens. 
\subsection{Chain of Multi-modal Thought}
\label{text and image}
Unlike the traditional MCoT,  
Chain of Multi-modal Thought (\datasetname{}) incorporates visual thought into rationale generation.
Formally, as shown in Figure~\ref{fig:background_comparison} (b), given an multi-modal input $\mathcal{X}_{inp}$, the model generates a multi-modal rationale step $\mathcal{R}^{m}_i$, which can be defined as:
\begin{equation}
	\small \mathcal{R}^{m}_{i} = \begin{cases} 
		\underset{\mathcal{R}^{vis}}{\operatorname{argmax}}\  \hat{\pi}(\mathcal{R}^{vis}|\mathcal{X}_{inp}, \mathbf{R}^{m}_{<i}), &\!\!\!\text{if } \hat{\pi}\! \geq \!\pi \\
		\underset{\mathcal{R}^{txt}}{\operatorname{argmax}}\  \pi(\mathcal{R}^{txt}|\mathcal{X}_{inp}, \mathbf{R}^{m}_{<i}), &\!\!\!\text{if } \hat{\pi}\! < \!\pi
	\end{cases}
\end{equation}
where $\hat{\pi}(\cdot)$ represents the probability that the model generates a rationale step with visual information, such as images or detailed descriptions of visual concepts.

\section{Quality Assurance}
We adopt Onboarding Test and Human Recheck method to ensure the quality of annotated data.

\noindent \textbf{Onboarding Test } All annotators must complete a preliminary test involving the annotation of 100 samples. Their annotations are assessed by three experts, and only those who achieve an accuracy of at least 85\% are allowed to continue to subsequent annotation tasks.

\noindent \textbf{Human Recheck } Following the onboarding test, annotators are required to recheck all data twice. This step ensures that each sample meets the multi-modal thought criteria and possesses coherent logical rationale. Only samples in \datasetname{} for which at least two annotators agree are accepted. The kappa coefficient among annotators reaches 0.93, indicating perfect agreement~\citep{landis1977measurement}.

\section{Annotation Details}
\label{appendix:annotation}

\subsection{Template Definition}
To standardize the format of all data in the \datasetname{} dataset, we design a multiple-choice question-answering template for all 4 tasks. The template includes five keys: \textit{question}, \textit{option}, \textit{image}, \textit{rationale}, and \textit{answer}. The specific content of the template is as follows:
\begin{tcolorbox}[colback=lightgray!20!white, colframe=lightgray!0!white, arc=2mm, width=\linewidth, boxrule=0.5pt]
	\textcolor{gray}{\textbf{\%Question\%}} \\
	\textit{[IMAGE0]} + \textit{$<$question$>$}
	\vspace{0.4\baselineskip}
	
	\textcolor{gray}{\textbf{\%Option\%}} \\
	(A) \textit{$<$option1$>$};\hspace{2em}(B) \textit{$<$option2$>$}; \\
	(C) \textit{$<$option3$>$};\hspace{2em}(D) \textit{$<$option4$>$}
	\vspace{0.4\baselineskip}
	
	\textcolor{gray}{\textbf{\%Image\%}} \\
	\textit{IMAGE0} :\hspace{1em}\textit{$<$image\_name-0$>$}, \\
	\textit{IMAGE1} :\hspace{1em}\textit{$<$image\_name-1$>$}, \\
	... ,\\ 
	\textit{IMAGEN} :\hspace{1em}\textit{$<$image\_name-N$>$}
	\vspace{0.4\baselineskip}
	
	\textcolor{gray}{\textbf{\%Rationale\%}} \\
	According to the question ...\textit{[IMAGE1]} ... And ... \textit{[IMAGEN]} ... Therefore, ...
	\vspace{0.4\baselineskip}
	
	\textcolor{gray}{\textbf{\%Answer\%}} \\
	A (the correct option)
	\vspace{0.4\baselineskip}
\end{tcolorbox}
The \textit{\%Question\%} consists of an image and text where the image is represented by an identifier \textit{[IMAGE0]}.
The \textit{\%Option\%} contains four similar options, only one of which is correct. The \textit{\%Image\%} section corresponds image identifiers to specific images. The \textit{\%Rationale\%} section consists of interleaved image identifiers and text. The \textit{\%Answer\%} section contains the correct answer option.

\subsection{Benchmark Annotation}
\subsubsection{Annotation Details}
We divide the datasets and distribute them to multiple annotators, who then complete and recheck all annotations. Additionally, we design some manual guidelines, which the annotators need to follow during the annotation process.

\subsubsection{Visual Creation}
The original GeoQA+~\citep{cao2022augmented} dataset provides geometric images along with textual question related to images. 

First, we extract the parts of the original dataset including questions, options and answers. Then, we split the answer section into reasoning part and final options. Afterwards, we reposition the image from input to the output rationale, and then modify the dataset format according to the template.

Note that we solely select samples where the question contains detailed and accurate descriptions of the image content, and then utilize "human recheck" to augment the question. Specifically, we supplement the descriptions based on the image content in the same language style. For example, we add descriptions such as \textit{"connect point A, B"} or \textit{"point A, C are located on both sides of diameter BD"}. 

The guideline instructions  are as follows:
\begin{tcolorbox}[colback=lightgray!20!white, colframe=lightgray!0!white, arc=2mm, width=\linewidth, boxrule=0.5pt]
	\textbf{[Instruction]}\\
	We will provide you with a geometric dataset and a template. Please follow the annotation flow outlined below:
	\begin{itemize}
		\item Extract subject, choices, and answer parts from the original dataset, separating the reasoning and final option.
		\item Reposition the image from input to the output rationale and modify the dataset to conform to the template format.
		\item Select samples whose text aligns with the image, and then supplement the descriptions within text according to the original dataset's linguistic style, based on the geometric shapes in the images.
	\end{itemize}
	\vspace{0.4\baselineskip}
\end{tcolorbox}

\subsubsection{Visual Deletion}
The original dataset JHU-CROWD++~\citep{sindagi2020jhu} provides images containing numerous faces and related boxing. 

First,  we extract samples whose face number falls within a specific range. According to the boxing, we mask 10 faces each time until all have been masked. We complete this process from the left side of image to right. Afterwards, we generate the textual components automatically to meet the template requirements

The guideline instructions  are as follows:
\begin{tcolorbox}[colback=lightgray!20!white, colframe=lightgray!0!white, arc=2mm, width=\linewidth, boxrule=0.5pt]
	\textbf{[Instruction]}\\
	We will provide you with a crowd-counting dataset and a template. Please follow the annotation flow outlined below:
	\begin{itemize}
		\item Sequentially box faces by setting them to be transparent according to the coordinates from left to right, boxing 10 faces at a time.
	\end{itemize}
	\vspace{0.4\baselineskip}
\end{tcolorbox}

\subsubsection{Visual Update}
The KILOGRAM~\citep{ji2022abstract} dataset provides tangrams along with related shape annotations. 

The original tangrams are stored within SVG files, which offers the possibility to  modify them. First, we color each label category and annotate them with arrows. After that, we save the SVG files as PNG images each time a label is added.
Subsequently, we filter out samples where significant overlaps exist within images or the label meaning isn't reasonable. Finally, we proceed with generating the textual components automatically. Note that we collect labels of the whole shape as the option pool.

The guideline instructions are as follows:
\begin{tcolorbox}[colback=lightgray!20!white, colframe=lightgray!0!white, arc=2mm, width=\linewidth, boxrule=0.5pt]
	\textbf{[Instruction]}\\
	We will provide you with a tangram recognition dataset and a template. Please follow the annotation flow outlined below:
	\begin{itemize}
		\item Draw different colors for regions of the tangram according to the annotations.
		\item Add arrows and labels to the tangram images according to the annotations within the dataset, adding one at a time and then saving the image.
	\end{itemize}
	\vspace{0.4\baselineskip}
\end{tcolorbox}

\subsubsection{Visual Selection}
The source data website\footnote{https://www.allstarpuzzles.com/spotdiff/index.html} provides pairs of "spot-diff" images along with the diff boxing.

Based on the boxing, we first crop out the diff from left to right and then place these cropped sections on a white background image of the same size as the original image, while maintaining their positions. Note that we control the number of diffs in each cropping to ensure that the generated images remain in four pairs.

Then, we horizontally concatenate the image pairs into a single one, using transparent gaps to separate the pairs. Then, we generate the text parts automatically.

The guideline instructions  are as follows:
\begin{tcolorbox}[colback=lightgray!20!white, colframe=lightgray!0!white, arc=2mm, width=\linewidth, boxrule=0.5pt]
	\textbf{[Instruction]}\\
	We will provide you with a spot-diff dataset and a template. Please follow the annotation flow outlined below:
	\begin{itemize}
		\item Extract the diffs according to the coordinates.
		\item Place diffs on a blank background equal in size to the original image step by step, ensuring that the new images contain 4 pairs and that the number of differences added each time is equal.
	\end{itemize}
	\vspace{0.4\baselineskip}
\end{tcolorbox}

\subsection{Distribution of Image Quality Scores}
\label{distribution of image scores}
We categorize the image quality scores into five levels, with scores ranging from 0 to 4, which from low to high respectively represent \textit{not Relevant at all}, \textit{Relevant but Logically Wrong}, \textit{Relevant and Logically Partially Correct}, \textit{Relevant and Logically Completely Correct}, \textit{Relevant and Logically Completely Correct and Beautiful}.

We define \textit{Relevant} as the generated image content being related to the topic. 
\begin{itemize}
	\item For \textbf {visual creation}, the generated image is considered relevant if it includes geometric shapes; 
	\item For \textbf{visual deletion}, the image is relevant if it includes a crowd; 
	\item For  \textbf{visual update}, the image is relevant if its content is similar to the corresponding tangram shapes; 
	\item For  \textbf{visual selection}, the image is relevant if the scene in the generated image matches the scene depicted in the input image.
\end{itemize}

We define \textit{Logically Correct} as the image content being consistent with the rationale generated by LVLMs. 
\begin{itemize}
	\item For  \textbf{visual creation}, the image is logically correct if the geometric content of the image matches the geometric description in the rationale; 
	\item For \textbf {visual deletion}, the image is logically correct if the number of people described in the rationale is similar to the number of people in the image or if the specific scene described in the rationale matches the image content; 
	\item For \textbf {visual update}, the image is logically correct if the objects described in the rationale are consistent with those displayed in the image;
	\item For \textbf {visual selection}, the image is logically correct if the scene described in the rationale matches the scene displayed in the image.
\end{itemize}

We randomly sample 50 instances that incorporate images within the rationale ( or all available instances if fewer than 50 ) for each of the four tasks for GILL, NExT-GPT, and AnyGPT respectively, ensuring an average distribution of the four prompt strategies as much as possible. Sampling is not performed in certain scenarios for methods like \texttt{Direct} that rarely produce rationales, thereby reducing the impact on judgment.

We select several annotators to complete the scoring task and provide them with manual guidelines as the scoring standard. Only scores that are agreed upon by at least three annotators are considered valid. The specific guideline instructions are as follows:
\begin{tcolorbox}[colback=lightgray!20!white, colframe=lightgray!0!white, arc=2mm, width=\linewidth, boxrule=0.5pt]
	\textbf{[Instruction$_{\texttt{1}}$]}\\
	We will provide you with some images and the associated rationale. Please score the image according to the following criteria:
	\begin{itemize}
	 \item 0 - Not relevant at all
	 \item 1 - Relevant but logically wrong
	 \item 2 - Relevant and logically partially correct
	 \item 3 - Relevant and logically completely correct
	 \item 4 - Relevant and logically completely correct and beautiful.
	\end{itemize}
\end{tcolorbox}

\begin{tcolorbox}[colback=lightgray!20!white, colframe=lightgray!0!white, arc=2mm, width=\linewidth, boxrule=0.5pt]
	\textbf{[Instruction$_{\texttt{2}}$]}\\
	Note the specific criteria for different tasks:\\
	\textit{(The specific criteria for Relevant and Logically Correct mentioned in above paragraph)}
	\vspace{0.4\baselineskip}
\end{tcolorbox}

\section{Experiment Details}
\label{appendix:experiment}

\subsection{Metrics}
Given that \datasetname{} is a multiple-choice question-answering dataset with fixed answers, we select \textit{accuracy} and \textit{Macro-F1} as the evaluation metrics for assessing model outputs.

\hspace*{\fill}

\subsection{Random Baseline}
We implement the random baseline by randomly selecting one from four options, and then abstract the average results with three attempts.

\hspace*{\fill}

\subsection{Prompting Strategy}
In addition to employing single-turn dialogue for obtaining answers in the \texttt{Direct} method, for the other three prompting strategies(\texttt{CoT}, \texttt{Desp-CoT} and \texttt{VoT}), we utilize a two-turn dialogue approach to have the model generate answers. 

In the first turn of the dialogue, we use designed prompts (details are in \ref{prompts design}) to prompt the model to generate reasoning. In the second turn of the dialogue, we prompt the model to select the final option through 'Therefore, among A through D, the answer is'. 

\hspace*{\fill}

\subsection{Prompts Design}
\label{prompts design}
For the four tasks in \datasetname{}, we design different prompt words. We follow the sequence of stating roles, outlining tasks, presenting specific questions, and finally supplementing various strategy words. 

Specifically, we design the \textit{State Roles} section to clarify the roles of LVLMs, thereby aiding LVLMs in focusing on the relevant domain knowledge. The \textit{Outline Task} section is designed to describe the task content and objectives, enhancing LVLMs' understanding of the intent. The \textit{Specific Question} section is designed to provide LVLMs with specific problems that need to be addressed. The \textit{Strategy Words} section is designed to implement different prompt strategies. 

The detailed prompt words for each task are as follows.

\hspace*{\fill}

\subsubsection{Visual Creation}
\phantom{placeholder}
\begin{tcolorbox}[colback=lightgray!20!white, colframe=lightgray!0!white, arc=2mm, width=\linewidth, boxrule=0.5pt]
	\textcolor{gray}{\textbf{\%State Roles\%}} \\
	As a math expert proficient in solving geometry problems, you will now face a geometry math question with four options.
	\vspace{0.4\baselineskip}
	
	\textcolor{gray}{\textbf{\%Outline Task\%}} \\
	You need to solve this geometry problem and select the correct answer from the given options.
	\vspace{0.4\baselineskip}
	
	\textcolor{gray}{\textbf{\%Specific Question\%}} \\
	\textit{$<$Question, Option$>$}
	\vspace{0.4\baselineskip}
\end{tcolorbox}

\begin{tcolorbox}[colback=lightgray!20!white, colframe=lightgray!0!white, arc=2mm, width=\linewidth, boxrule=0.5pt]
	\textcolor{gray}{\textbf{\%Strategy Words\%}} \\
	Among A through D, the answer is(\texttt{Direct}) / \\
	Let's think step by step.(\texttt{CoT}) / \\
	Describe the image information relevant to the question.(\texttt{Desp-CoT}) / \\
	Visualize the state after each reasoning step.(\texttt{VoT})
	\vspace{0.4\baselineskip}
\end{tcolorbox}

\subsubsection{Visual Deletion}
\phantom{placeholder}
\begin{tcolorbox}[colback=lightgray!20!white, colframe=lightgray!0!white, arc=2mm, width=\linewidth, boxrule=0.5pt]
	\textcolor{gray}{\textbf{\%State Roles\%}} \\
	You are currently playing the role of an expert skilled at accurately counting the number of people in an image.
	\vspace{0.4\baselineskip}
	
	\textcolor{gray}{\textbf{\%Outline Task\%}} \\
	Below, you will receive an image, a question, and four options. You need to answer the question and choose the correct answer from the provided options.
	\vspace{0.4\baselineskip}
	
	\textcolor{gray}{\textbf{\%Specific Question\%}} \\
	\textit{$<$Image, Question, Option$>$}
	\vspace{0.4\baselineskip}
\end{tcolorbox}

\begin{tcolorbox}[colback=lightgray!20!white, colframe=lightgray!0!white, arc=2mm, width=\linewidth, boxrule=0.5pt]
	\textcolor{gray}{\textbf{\%Strategy Words\%}} \\
	Among A through D, the answer is(\texttt{Direct}) / \\
	Let's think step by step.(\texttt{CoT}) / \\
	Describe the image information relevant to the question.(\texttt{Desp-CoT}) / \\
	Visualize the state after each reasoning step.(\texttt{VoT})
	\vspace{0.4\baselineskip}
\end{tcolorbox}

\subsubsection{Visual Update}
\phantom{placeholder}
\begin{tcolorbox}[colback=lightgray!20!white, colframe=lightgray!0!white, arc=2mm, width=\linewidth, boxrule=0.5pt]
	\textcolor{gray}{\textbf{\%State Roles\%}} \\
	You are now an expert in identifying tangram shapes.
	\vspace{0.4\baselineskip}
	
	\textcolor{gray}{\textbf{\%Outline Task\%}} \\
	Below, you will receive an image, a question, and four options. You need to answer the question and choose the correct answer from the provided options.
	\vspace{0.4\baselineskip}
	
	\textcolor{gray}{\textbf{\%Specific Question\%}} \\
	\textit{$<$Image, Question, Option$>$}
	\vspace{0.4\baselineskip}
\end{tcolorbox}

\begin{tcolorbox}[colback=lightgray!20!white, colframe=lightgray!0!white, arc=2mm, width=\linewidth, boxrule=0.5pt]
	\textcolor{gray}{\textbf{\%Strategy Words\%}} \\
	Among A through D, the answer is(\texttt{Direct}) / \\
	Let's think step by step.(\texttt{CoT}) / \\
	Describe the image information relevant to the question.(\texttt{Desp-CoT}) / \\
	Visualize the state after each reasoning step.(\texttt{VoT})
	\vspace{0.4\baselineskip}
\end{tcolorbox}

\subsubsection{Visual Selection}
\phantom{placeholder}
\begin{tcolorbox}[colback=lightgray!20!white, colframe=lightgray!0!white, arc=2mm, width=\linewidth, boxrule=0.5pt]
	\textcolor{gray}{\textbf{\%State Roles\%}} \\
	You are now an expert in spot the difference games.
	\vspace{0.4\baselineskip}
	
	\textcolor{gray}{\textbf{\%Outline Task\%}} \\
	Below, you will receive an image, a question, and four options. You need to answer the question and choose the correct answer from the provided options.
	\vspace{0.4\baselineskip}
	
	\textcolor{gray}{\textbf{\%Specific Question\%}} \\
	\textit{$<$Image, Question, Option$>$}
	\vspace{0.4\baselineskip}
\end{tcolorbox}

\begin{tcolorbox}[colback=lightgray!20!white, colframe=lightgray!0!white, arc=2mm, width=\linewidth, boxrule=0.5pt]
	\textcolor{gray}{\textbf{\%Strategy Words\%}} \\
	Among A through D, the answer is(\texttt{Direct}) / \\
	Let's think step by step.(\texttt{CoT}) / \\
	Describe the image information relevant to the question.(\texttt{Desp-CoT}) / \\
	Visualize the state after each reasoning step.(\texttt{VoT})
	\vspace{0.4\baselineskip}
\end{tcolorbox}

\section{Analysis Details}
\label{appendix:analysis}
\subsection{ROSCOE}
For ROSCOE~\citep{golovneva2023roscoe} computation, we first calculate the average score for data from the four tasks in \datasetname{} separately and then average these four individual scores to obtain a final average.

\subsection{CLIPScore}
For CLIPScore~\citep{hessel2021clipscore} computation, we first calculate the score for each sample in the task by comparing every image in the reasoning path constructed in \datasetname{} with the rationale generated by LVLMs, and then averaging the scores from multiple images as the result of this sample. Subsequently, we calculate the average score of each task.

\subsection{In-context Learning Details}
For valid in-context learning, we annotate 10 additional pieces of data for each task as the development set in the same way as test. The dev set covers all options evenly. In addition, during the experiment, $k$ items were randomly sampled as the demonstrations of $k$-shot.

Following the work of \citet{chen2024m}, we used the following prompt template for multi-modal in-context-learning:
\begin{tcolorbox}[colback=lightgray!20!white, colframe=lightgray!0!white, arc=2mm, width=\linewidth, boxrule=0.5pt]
	\textcolor{gray}{\textbf{\%Example 1\%}} \\
	\textbf{[QUESTION]} \\
	\textit{$<$Question, Image$>$}
	\vspace{0.4\baselineskip}
	
	\textbf{[OPTION]} \\
	\textit{(A) $<$Option1$>$; (B) $<$Option2$>$; ...}
	\vspace{0.4\baselineskip}
	
	\textbf{[RATIONALE]} \\
	\textit{$<$Rationale$>$}
	\vspace{0.4\baselineskip}
	
	\textbf{[Answer]} \\
	\textit{$<$Answer$>$}
	\vspace{0.4\baselineskip}
	
	\textcolor{gray}{\textbf{\%Example 2\%}}... \\
	
	\textcolor{gray}{\textbf{\%Example N\%}} \\
	\textbf{[QUESTION]} \\
	\textit{$<$Question, Image$>$}
	\vspace{0.4\baselineskip}
	
	\textbf{[OPTION]} \\
	\textit{(A) $<$Option1$>$; (B) $<$Option2$>$; ...}
	\vspace{0.4\baselineskip}
	
\end{tcolorbox}
Furthermore, for multi-modal output, rationale is an interleaved list of images and texts. Single-modal rationale only contains the text content.

\renewcommand{\thefigure}{10}
\begin{figure}[t]
	\centering
	\includegraphics[width=\linewidth]{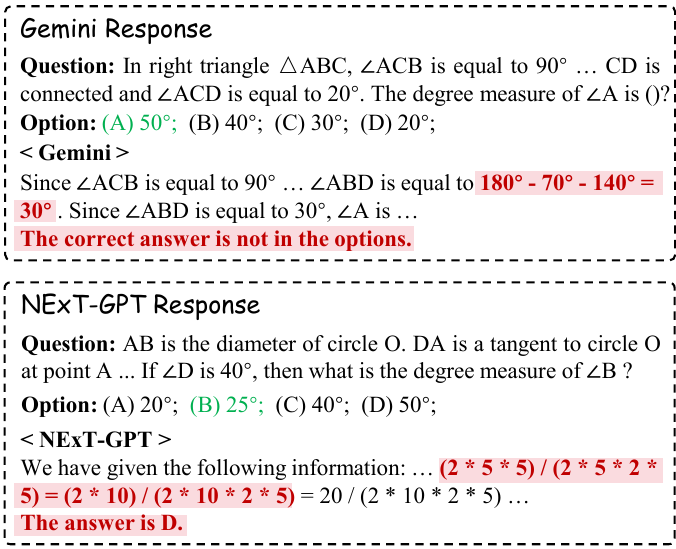}
	\caption{Logical errors in textual statements for \textit{Gemini-Pro}~\citep{team2023gemini} and \textit{NExT-GPT}~\citep{wu24next}.}
	\label{fig:text-quality-casestudy}
\end{figure}

\label{appendix:more_results}

\section{Complete Experiment Results}

Complete evaluation results of LVLMs on \datasetname{}, as shown in Table \ref{table:more_results}.

\section{Irrelevant Image and Text Logic}

Effective vision logic is crucial for visual reasoning. However, we observe that LVLMs sometimes generate irrelevant text and image logic across \datasetname{} tasks, which hinders the reasoning process. This highlights the challenges current LVLMs face in integrating effective visual logic for visual reasoning. Specific examples are as follows.
\subsection{Accurate Text with Inaccurate Image}
During the experiments, we observe cases where LVLMs generate accurate textual reasoning but produce images irrelevant to the problem. As shown in Figure \ref{fig:casestudy}(a), NExT-GPT~\cite{wu2023nextgpt} are expected to generate an image that contains geometric shapes consistent with the rationale. However, due to the lack of effective visual logic, LVLMs produce images that only contain textual content, which does not aid in promoting visual reasoning.

\subsection{Accurate Image with Inaccurate Text}
There are cases where LVLMs generate accurate images according to task requirements but produce text descriptions inconsistent with these images. As shown in Figure \ref{fig:casestudy}(b), AnyGPT~\cite{zhan2024anygpt} generate an image of a fox consistent with the correct answer; however, the rationale determines the answer to be a beagle. This reflects the current LVLMs struggle to perform further reasoning based on the generated images, indicating a lack of effective visual logic.


\section{Ethical Considerations}

\noindent \textbf{Data Access} We collect data from GeoQA+~\citep{cao2022augmented}, JHU-CROWD++ dataset~\citep{sindagi2020jhu}, KILOGRAM~\citep{ji2022abstract} and online websites\footnote{https://www.allstarpuzzles.com/spotdiff/index.html}. These datasets are all open-source and permitted for academic research, complying with ethical commitments for data usage.

\renewcommand{\thefigure}{11}
\begin{figure}[t]
	\centering
	\subfigure[Accurate text with inaccurate image]{
		\begin{minipage}[b]{0.9\linewidth}
			\includegraphics[width=\linewidth]{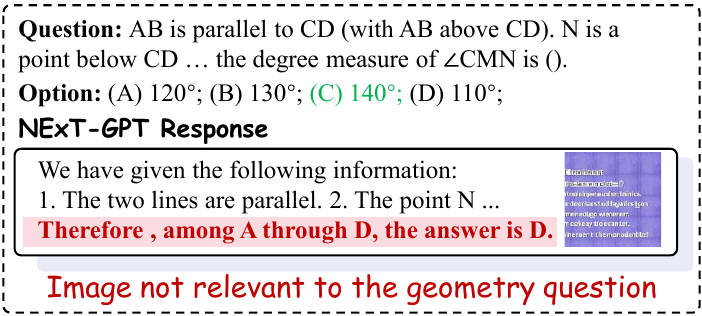} 
		\end{minipage}
	}
	\subfigure[Accurate image with inaccurate text]{
		\begin{minipage}[b]{0.9\linewidth}
			\includegraphics[width=\linewidth]{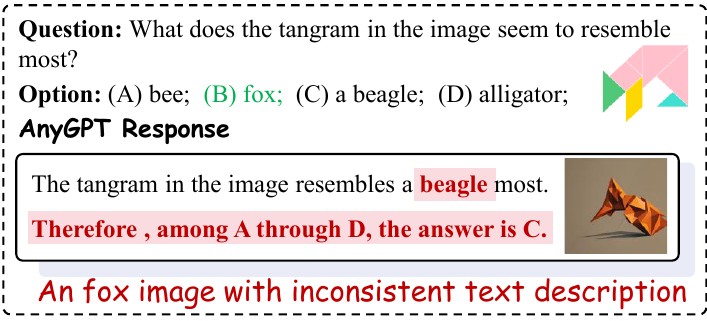} 
		\end{minipage}
	}
	\caption{Inconsistency between text and images within the rationale output by LVLMs} 
	\label{fig:casestudy}
\end{figure}

\renewcommand{\thetable}{4}
\begin{table*}[t]
	\centering
	\begin{adjustbox}{width=0.99\textwidth}
		\begin{tabular}{lcccccccccc}
			\toprule
			\multirow{2}{*}{Model} & \multicolumn{2}{c}{Visual Creation} & \multicolumn{2}{c}{Visual Deletion} & \multicolumn{2}{c}{Visual Update} & \multicolumn{2}{c}{Visual Selection} & \multicolumn{2}{c}{Average}
			\\\cmidrule{2-11}
			& Acc & Macro-F1 & Acc & Macro-F1 & Acc & Macro-F1 & Acc & Macro-F1 & Acc & Macro-F1
			\\
			\midrule
			Random & 27.10 & 26.75 & 25.17 & 25.15 & 24.06 & 24.05 & 25.59 & 25.55 & 25.48 & 25.37
			\\
			\midrule
			\rowcolor{gray!8}\multicolumn{11}{c}{\textit{Qwen-VL-7B}~\cite{Qwen-VL}}\\
			\midrule
			\texttt{Direct}~\cite{Qwen-VL} & 21.49 & 12.78 & \underline{26.35} & \underline{18.29} & \underline{37.64} & \underline{30.34} & 22.08 & 13.80 & \underline{26.89} & \underline{18.80} \\
			\texttt{CoT}~\cite{kojima2022large} & \underline{23.96} & \underline{19.22} & 12.63 & 11.81 & 33.62 & 26.13 & 23.22 & \underline{18.00} & 23.26 & 18.79 \\
			\texttt{Desp-CoT}~\cite{wu2023role} & 19.90 & 13.23 & 20.94 & 7.73 & 30.59 & 23.85 & \underline{26.05} & 10.48 & 24.37 & 13.82 \\ 
			\texttt{VoT}~\cite{wu2024visualization} & 22.08 & 17.51 & 14.43 & 11.71 & 28.52 & 21.02 & 22.08 & 12.47 & 21.78 & 15.68 \\ 
			\midrule
			\rowcolor{gray!8}\multicolumn{11}{c}{\textit{LLaVA-NeXT-13B}~\cite{liu2024llavanext}}\\
			\midrule
			\texttt{Direct}~\cite{liu2024llavanext} & \underline{26.34} & \underline{19.72} & 20.64 & \underline{20.06} & \underline{35.47} & \underline{34.26} & 22.76 & 19.60 & \underline{26.30} & \underline{23.41} \\
			\texttt{CoT}~\cite{kojima2022large} & 22.18 & 12.33 & 21.44 & 15.21 & 26.36 & 18.99 & 24.92 & 19.91 & 23.73 & 16.61 \\
			\texttt{Desp-CoT}~\cite{wu2023role} & 19.90 & 12.82 & 23.45 & 17.47 & 27.01 & 18.82 & 25.59 & \underline{20.77} & 23.99 & 17.47 \\ 
			\texttt{VoT}~\cite{wu2024visualization} & 20.79 & 15.58 & \underline{25.55} & 18.55 & 27.55 & 18.95 & \underline{26.61} & 17.23 & 25.13 & 17.58 \\ 
			\midrule
			\rowcolor{gray!8}\multicolumn{11}{c}{\textit{GILL}~\cite{koh2023generating}}\\
			\midrule
			\texttt{Direct}~\cite{koh2023generating} & \underline{16.93} & \underline{15.75} & \underline{22.6}5 & \underline{13.90} & \underline{23.43} & \underline{12.62} & 18.12 & \underline{10.16} & \underline{20.28} & \underline{13.11} \\
			\texttt{CoT}~\cite{kojima2022large} & 8.61 & 9.96 & 12.63 & 8.62 & 18.11 & 8.20 & 17.21 & 8.34 & 14.14 & 8.78 \\
			\texttt{Desp-CoT}~\cite{wu2023role} & 6.83 & 7.93 & 20.74 & 9.60 & 21.69 & 10.90 & \underline{20.95} & 9.12 & 17.55 & 9.39 \\ 
			\texttt{VoT}~\cite{wu2024visualization} & 5.94 & 7.01 & 17.94 & 11.81 & 21.04 & 11.51 & 14.27 & 9.23 & 14.80 & 9.89 
			\\
			\midrule
			\rowcolor{gray!8}\multicolumn{11}{c}{\textit{NExT-GPT}~\cite{wu24next}}\\
			\midrule
			\texttt{Direct}~\cite{wu2023nextgpt} & \underline{24.26} & \underline{19.00} & \underline{25.75} & \underline{19.15} & 24.30 & \underline{18.04} & 22.42 & 16.24 & \underline{24.18}  & \underline{18.11} \\
			\texttt{CoT}~\cite{kojima2022large} & 20.20 & 13.88 & 23.85 & 17.25 & 22.78 & 17.95 & 21.52 & \underline{18.39} & 22.09 & 16.87 \\
			\texttt{Desp-CoT}~\cite{wu2023role} & 17.52 & 13.93 & 23.95 & 14.13 & \underline{25.38} & 17.91 & \underline{22.99} & 16.90 & 22.46 & 15.72 \\ 
			\texttt{VoT}~\cite{wu2024visualization} & 13.17 & 12.91 & 22.85 & 14.38 & 25.05 & 16.28 & 22.88 & 18.32 & 20.99 & {15.47} \\
			\midrule
			\rowcolor{gray!8}\multicolumn{11}{c}{\textit{AnyGPT}~\cite{zhan2024anygpt}}\\
			\midrule
			\texttt{Direct}~\cite{zhan2024anygpt} & 19.11 & 12.18 & 17.43 & 11.92 & 23.10 & 17.85 & \underline{27.63} & 16.91 & 21.82 & 14.72 \\
			\texttt{CoT}~\cite{kojima2022large} & 10.10 & 10.36 & 21.74 & 11.96 & 24.08 & 18.37 & 22.20 & 15.77 & 19.53 & 14.12 \\
			\texttt{Desp-CoT}~\cite{wu2023role} & \underline{19.31} & \underline{14.15} & 22.75 & \underline{12.22} & 24.84 & 18.72 & 25.59 & 16.63 & \underline{23.12} & \underline{15.43} \\  
			\texttt{VoT}~\cite{wu2024visualization} & 11.78 & 10.22 & \underline{23.45} & 11.45 & \underline{26.36} & \underline{19.44} & 25.59 & \underline{18.43} & 21.80 & 14.89 \\
			\midrule
			\rowcolor{gray!8}\multicolumn{11}{c}{\textit{Gemini}~\cite{team2023gemini}}\\
			\midrule
			\texttt{Direct}~\cite{team2023gemini} & 28.91 & 25.43 & \textbf{\underline{30.86}} & 22.28 & \underline{46.36} & \underline{46.26} & \textbf{\underline{27.63}} & 20.69 & \textbf{\underline{33.44}} & \textbf{\underline{28.67}} \\
			\texttt{CoT}~\cite{kojima2022large} & 27.92 & 23.07 & 28.76 & \underline{22.73} & 40.24 & 40.02 & 27.39 & \textbf{\underline{23.60}} & 31.08 & 27.36 \\
			\texttt{Desp-CoT}~\cite{wu2023role} & 18.04 & 14.61 & 29.36 & 21.43 & 31.05 & 23.20 & 25.14 & 11.32 & 25.90 & 17.64 \\ 
			\texttt{VoT}~\cite{wu2024visualization} & \textbf{\underline{33.27}} & \textbf{\underline{26.48}} & 27.05 & 20.79 & 35.36 & 27.83 & 24.92 & 19.38 & 30.15 & 23.62 \\
			\midrule
			\rowcolor{gray!8}\multicolumn{11}{c}{\textit{DeepSeek-VL-7B}~\cite{lu2024deepseekvl}}\\
			\midrule
			\texttt{Direct}~\cite{lu2024deepseekvl} & 25.15 & 19.97 & 24.25 & 17.18 & \textbf{\underline{47.51}} & \textbf{\underline{47.12}} & 25.71 & 14.71 & \underline{30.66} & \underline{24.75} \\
			\texttt{CoT}~\cite{kojima2022large} & 25.94 & 20.03 & \underline{26.35} & \underline{20.39} & 42.73 & 34.35 & 23.56 & 15.72 & 29.65 & 22.62 \\
			\texttt{Desp-CoT}~\cite{wu2023role} & 20.00 & 15.95 & 23.95 & 10.82 & 35.03 & 26.93 & \underline{27.41} & 9.54 & 26.60 & 15.81 \\ 
			\texttt{VoT}~\cite{wu2024visualization} & \underline{26.63} & \underline{20.53} & 18.54 & 15.17 & 45.44 & 36.43 & 23.90 & \underline{16.19} & 28.63 & 22.08 \\ 
			\midrule		
			\rowcolor{gray!8}\multicolumn{11}{c}{\textit{IDEFICS2-8B}~\cite{laurenccon2024matters}}\\
			\midrule
			\texttt{Direct}~\cite{laurenccon2024matters} & \underline{32.97} & \underline{22.02} & 26.65 & 15.98 & \underline{41.11} & \underline{40.30} & 24.80 & 10.35 & \underline{31.38} & \underline{22.16} \\
			\texttt{CoT}~\cite{kojima2022large} & 24.46 & 19.58 & 28.96 & 22.22 & 26.90 & 17.40 & 22.31 & 14.90 & 25.66 & 18.53 \\
			\texttt{Desp-CoT}~\cite{wu2023role} & 22.38 & 14.91 & 23.85 & 12.34 & 31.67  & 28.97 & \underline{27.63} & 14.65 & 26.38 & 17.72 \\ 
			\texttt{VoT}~\cite{wu2024visualization} & 24.46 & 19.62 & \underline{29.96} & \textbf{\underline{22.77}} & 29.83 & 25.73 & 23.90 & \underline{15.92} & 27.04 & 21.01 \\
			\midrule
			\rowcolor{gray!8}\multicolumn{11}{c}{\textit{InstructBlip-13B}~\cite{instructblip}}\\
			\midrule
			\texttt{Direct}~\cite{instructblip} & \underline{23.27} & \underline{18.20} & \underline{26.25} & \underline{13.95} & 29.72 & 17.38 & \underline{22.88} & \underline{15.19} & \underline{25.53} & \underline{16.18} \\
			\texttt{CoT}~\cite{kojima2022large} & 21.09 & 15.39 & 18.84 & 11.90 & \underline{33.41} & \underline{26.17} & 11.55 & 9.25 & 21.22 & 15.68 \\
			\texttt{Desp-CoT}~\cite{wu2023role} & 11.29 & 6.20 & 12.22 & 6.54 & 11.93 & 9.07 & 10.31 & 6.27 & 11.44 & 7.02 \\ 
			\texttt{VoT}~\cite{wu2024visualization} & 1.98 & 2.98 & 3.71 & 4.93 & 18.98 & 17.54 & 1.02 & 1.54 & 6.42 & 6.75 \\ 
			
			\bottomrule
		\end{tabular}
	\end{adjustbox}
	\caption{
		The complete results on various LVLMs. The \textbf{bold content} indicates the best performance across all models and all methods, while the \underline{underlined content} signifies the best performance within a single model across all methods.
	}
	\label{table:more_results}
\end{table*}

\noindent \textbf{Participant Recruitment} We recruit participants from multiple universities and require each participant to meet a language proficiency requirement of either passing the CET-6 exam or scoring 6 or above on the IELTS. Additionally, all participants are from various regions, which may introduce some regional biases. We constrain the dataset to common human knowledge to minimize national differences. All annotators have signed informed consent files and receive compensation above the local minimum wage standards. Furthermore, this study does not need IRB review.

\noindent \textbf{Dataset Collection Process} Our annotation process requires participants to first pass a test with 100 example questions. During this phase, participants receive a compensation of \$15 aimed at familiarizing them with the task. Subsequently, annotators are paid \$10 per hour, totaling approximately 300 human-hours for manual annotation. Additionally, an extra 40 hours are allocated for rechecking to ensure accurate annotation. Overall, we employ five experts and three students to complete the annotation and rechecking processes.

\end{document}